\documentclass[conference]{IEEEtran}
\IEEEoverridecommandlockouts
\usepackage{cite}
\usepackage{amsmath,amssymb,amsfonts}
\usepackage{algorithmic}
\usepackage{graphicx}
\usepackage{textcomp}
\usepackage{xcolor}
\usepackage{booktabs}
\usepackage{multirow}
\def\BibTeX{{\rm B\kern-.05em{\sc i\kern-.025em b}\kern-.08em
    T\kern-.1667em\lower.7ex\hbox{E}\kern-.125emX}}
    
\usepackage[binary-units]{siunitx}
\usepackage[labelformat=simple,font=footnotesize,labelsep = period]{subcaption}
\usepackage{cuted}
\usepackage{stfloats}

\begin{document}

\thispagestyle{empty}
\newpage
\onecolumn
\begin{center}
This paper has been accepted for publication in 2020 IEEE OES Autonomous Underwater Vehicle (AUV) Symposium.
\vspace{0.75cm}\\
DOI: \\ 
IEEE Xplore: \\
\vspace{1.25cm}
\end{center}
©2020 the authors under a Creative Commons Licence CC-BY-NC-ND. Personal use of this material is permitted. Permission from IEEE must be obtained for all other uses, in any current or future media, including reprinting/republishing this material for advertising or promotional purposes, creating new collective works, for resale or redistribution to servers or lists, or reuse of any copyrighted component of this work in other works.
\twocolumn

\title{SeaShark: Towards a Modular Multi-Purpose Man-Portable AUV 
}

\author{\IEEEauthorblockN{Jesper Haahr Christensen \hspace{0.5cm} Marco Jacobi \hspace{0.5cm} Martin Clemmensen Rotne \\ 
Morten Soede Nielsen  \hspace{0.5cm} Max Abildgaard \hspace{0.5cm} Claus Eriksen \hspace{0.5cm} Lars Valdemar Mogensen}
\IEEEauthorblockA{\textit{ATLAS MARIDAN, 2960 Rungsted Kyst, Denmark} \\
{\tt\small \{jhc,mja,mcr,msn,mab,cer,lvm\}@atlasmaridan.com}
}}

\maketitle
\begin{figure*}[b!]
    \centering
    \includegraphics[width=1\linewidth]{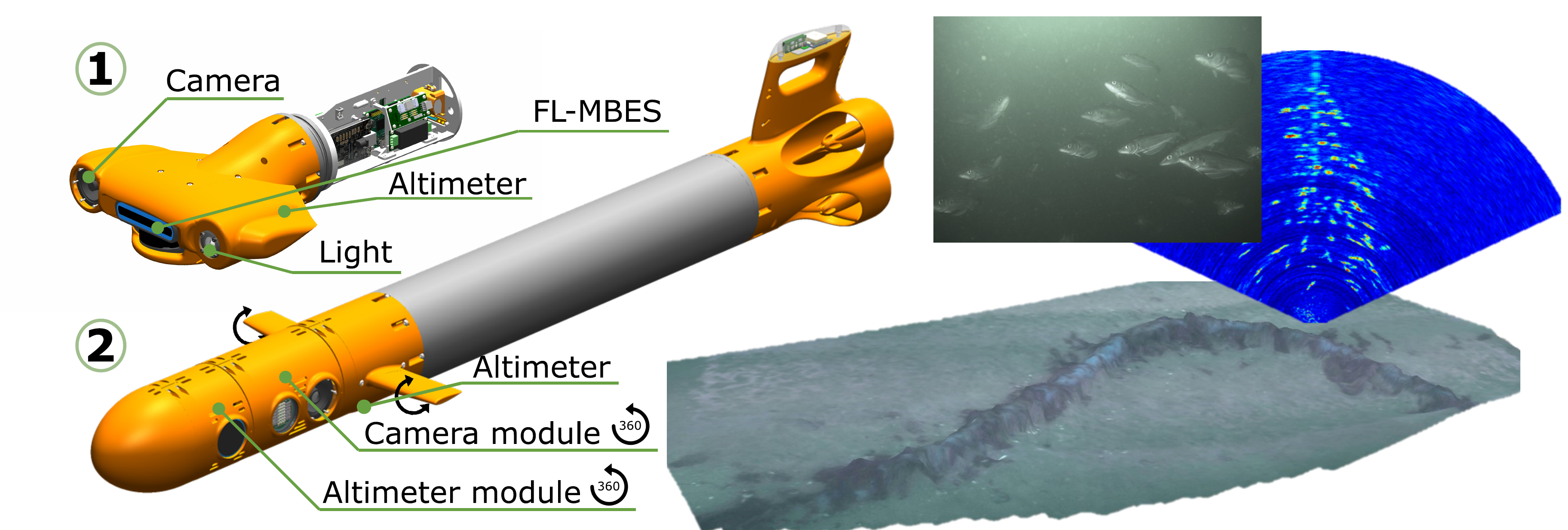}
    \caption{\textit{Overview of our modular multi-purpose AUV systems.} \textbf{1)} Exchangeable payload head with a forward-looking multibeam echosounder, camera, and light. This configuration enables time-synchronized correspondence of optical images and sonar images. \textbf{2)} Modular payload head with \ang{360} rotatable add-on sections. This configuration enables multi-view camera images, multi-point structure tracking, and modularity to add/remove/rotate sections to accommodate many objectives. }
    \label{fig:vehicle}
\end{figure*}

\begin{abstract}
In this work, we present the SeaShark AUV: a modular, easily configurable, one-man portable {\textmu}-AUV. The SeaShark AUV is conceived as modular parts that fit around a central main tube, which holds battery and other vital parts. The head unit comprises easy exhangeable, stackable, and 360\textbf{\textdegree} rotatable payload sections to quickly obtain a suitable configuration for many objectives. We employ navigation no better than dead reckoning or relative navigation with respect to some well-known structure, and thus aim at underwater activities that do not require highly accurate geo-referenced data-points. Operating the SeaShark AUV requires only the vehicle itself and a tablet for mission planning and post-mission review. We have built several complete SeaShark systems and have begun exploring the many possibilities and use-cases in both research and commercial use. Here we present a comprehensive overview and introduction to our AUV and operation principles, and further show data examples for experimental operations for shore-to-sea bio-habitat mapping and in-harbor wall and pier inspection.
\end{abstract}


\section{Introduction}
\label{sec:intro}
With oceans covering approximately \SI{72}{\percent} of our planet, and with as little as \SI{0.05}{\percent} known in meter resolution~\cite{knownoecean}, we continuously seek to explore unknown territories and provide better maps of what is already known. With modern technology, submerged robots is a popular tool favoring reduced risk, time, and operating cost. State-of-the-art autonomous underwater vehicles (AUVs) are capable of surveying ocean floors with high accuracy and long endurance~\cite{Kalwa2012,toytoauv}. 
In addition, and as an alternative, to high-cost survey-grade AUVs, low-cost remotely operated vehicles (ROVs) exist. These are to be manually piloted for off-shore inspection operations. However, this has clear limitations in that manual labor and tether management (length and control) are required. Bridging the gap between survey-grade AUVs and low-cost ROVs, we arrive at a smaller, less accurate, and less costly AUV. This still posses the agility and automation potential of the survey-grade AUVs and has a size, portability, and operational ease similar to that of ROVs. We argue that many underwater activities do not require navigational accuracy any better than dead reckoning or relative navigation with respect to some well-known structure. Additionally, we find that many tasks may be solved with a minimum of low-cost payloads typically not offered separately or in a configuration where it can be applied to a broad range of tasks.

In this paper, we present the SeaShark AUV. This aims at a novel one-man-portable, multi-purpose, and easily modifiable AUV system with many applications in tasks where highly accurate geo-referenced data-points are not a crucial element. As shown in Fig.~\ref{fig:vehicle}, the AUV consist of a base unit comprising all essential modules ranging from battery management, communication to depth and altitude sensors. On top of our base unit, we introduce modular sections, each comprising a single payload sensor. These can all be stacked and rotated \ang{360}, deeming it possible to obtain a payload configuration designed explicitly to each mission objective within minutes. With these options, we envision use-cases within difficult tasks such as tunnel inspections, aquatic vegetation monitoring, multi-view optical image geometry, synchronized optical and acoustic measurements. 
For example, relative navigation with multiple constraints on vehicle placement, e.g., wall following within a tunnel, harbor, or tank while maintaining an altitude reference to the bottom, is easily obtained by using multiple altimeters and orienting these to satisfy both constraints.

We have built several complete SeaShark AUV systems and are continuously exploring the many research and commercial use possibilities. 
 
 
\section{Related Work}
The authors were involved in several projects to develop and operate AUVs and subsea equipment in different sizes for many different operational objectives and applications. These applications involve underwater surveys, mining, water sampling, and environmental monitoring.

Gathering data from the underwater environment requires dedicated methods and tools. Some of these need user interaction, while others are automated when released. \\

\noindent\textbf{Sampler, profilers and towfish systems} \quad 
are classical underwater survey and measurement tools. Examples are dropcams, CTD profilers, towed cams, and towfish side-scan sonars. These systems are mainly deployed from boats of different sizes and need to be supervised by an operator. They deliver continuous data, which can be monitored in real-time, and data acquisition parameters are easily adapted during operation. \\

\noindent\textbf{Moored systems} \quad 
are stationary and can be used for long term data acquisition where a fixed measurement location is sufficient or required. 
Examples for such systems are moored water profilers with winch, observation structures installed on the seafloor, measurement equipment mounted on offshore windmills or oil rigs. \\

\noindent\textbf{Remotely Operated Vehicles (ROVs)} \quad 
are unmanned tethered systems that can move underwater and execute different tasks like inspection and manipulation. Different ROV classes exist: Heavyweight work class ROVs used mainly in the oil and gas industry, midsize interception and observation class ROVs, and finally small and lightweight mini-ROVs. In the smallest class, the market entry of the  BlueROV~\cite{BlueROV} enabled many new applications and research fields due to its availability and open architecture. It can be modified in an easy manner to adapt to new tasks. \\

\noindent\textbf{Autonomous Underwater Vehicles (AUVs)} \quad 
are unmanned underwater vehicles (UUVs) that operate freely underwater. Several systems are developed within the ATLAS group and deployed for different tasks such as underwater surveys (Maridan M600~\cite{wille2005sound}), underwater minesweeping (SeaOtter~\cite{ATLAS}), harbor and ship hull inspection (SeaCat and CView~\cite{Kalwa2012,Jacobi.2014b}), cable and pipeline inspection (KAPITAS AUV \cite{Jacobi.2014b}), environmental monitoring \cite{Eichhorn.2018} or gas flare sampling (IMGAM AUV \cite{Abildgaard2017}). 
All these AUVs are highly automated systems that operate autonomously with a pre-programmed mission plan. The vehicles with inspection and sampling capabilities have improved autonomy functions; they can leave the pre-programmed mission and execute their inspection and sampling tasks based on sensor measurements. \\

\begin{figure}[t]
    \centering
    \includegraphics[width=0.95\linewidth]{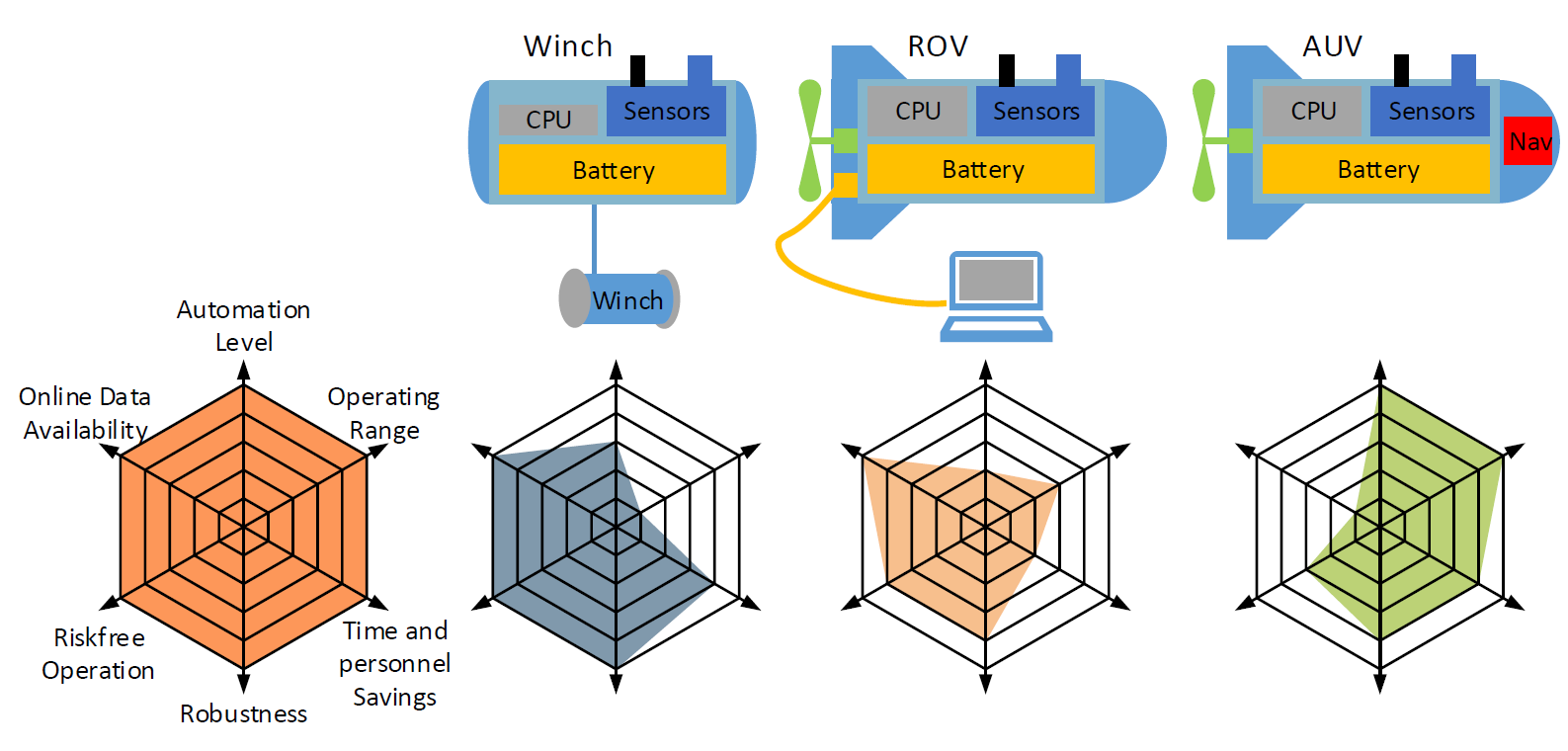}
    \caption{\textit{Comparison of different underwater monitoring systems \cite{Eichhorn.2018}.} The compared systems from left to right are towed systems (winch), ROV systems, and AUV systems.}
    \label{fig:uwsystems}
\end{figure}

All of the above have their tasks in underwater applications with each of their benefits and constraints. With ROVs and other tethered systems, real-time data is available and can be observed by the users. However, the operational range of these systems is limited when larger areas need to be surveyed. This is a clear advantage of an AUV. Its operational range is extensive and not limited by tether length or skilled in-mission personal. Fig.~\ref{fig:uwsystems} gives a short overview of operational constraints of the different system classes.

Conclusively, given the experience obtained in different projects and applications, we aim to develop a small one-man-portable underwater vehicle with an easy-to-use approach, arming us with an additional tool in the toolbox for underwater surveys. The vehicle system design and usage is described in the following sections.

\section{Vehicle Overview}

As shown in Fig.~\ref{fig:vehicle}, the SeaShark AUV is comprised of a base unit (thruster section and tube) and a head unit. The base unit includes all components that are essential to operating the AUV. Sensors such as depth, temperature, altitude, IMU, GNSS, communication, are all included in the base unit. We currently have (built and tested) two configurations for the payload head. This is, as shown in Fig.~\ref{fig:vehicle}, 1) A fixed head comprising a forward-looking multibeam imaging sonar, camera system, and light. 2) A modular head comprising exchangeable, stackable, and rotatable payload sections. In our payload sections, we currently include a high-quality industrial camera, a proprietary \SI{25000}{} lumen flash unit, and a single-beam echosounder to sample water-column data.

\begin{figure}
    \centering
    \includegraphics[width=1\linewidth]{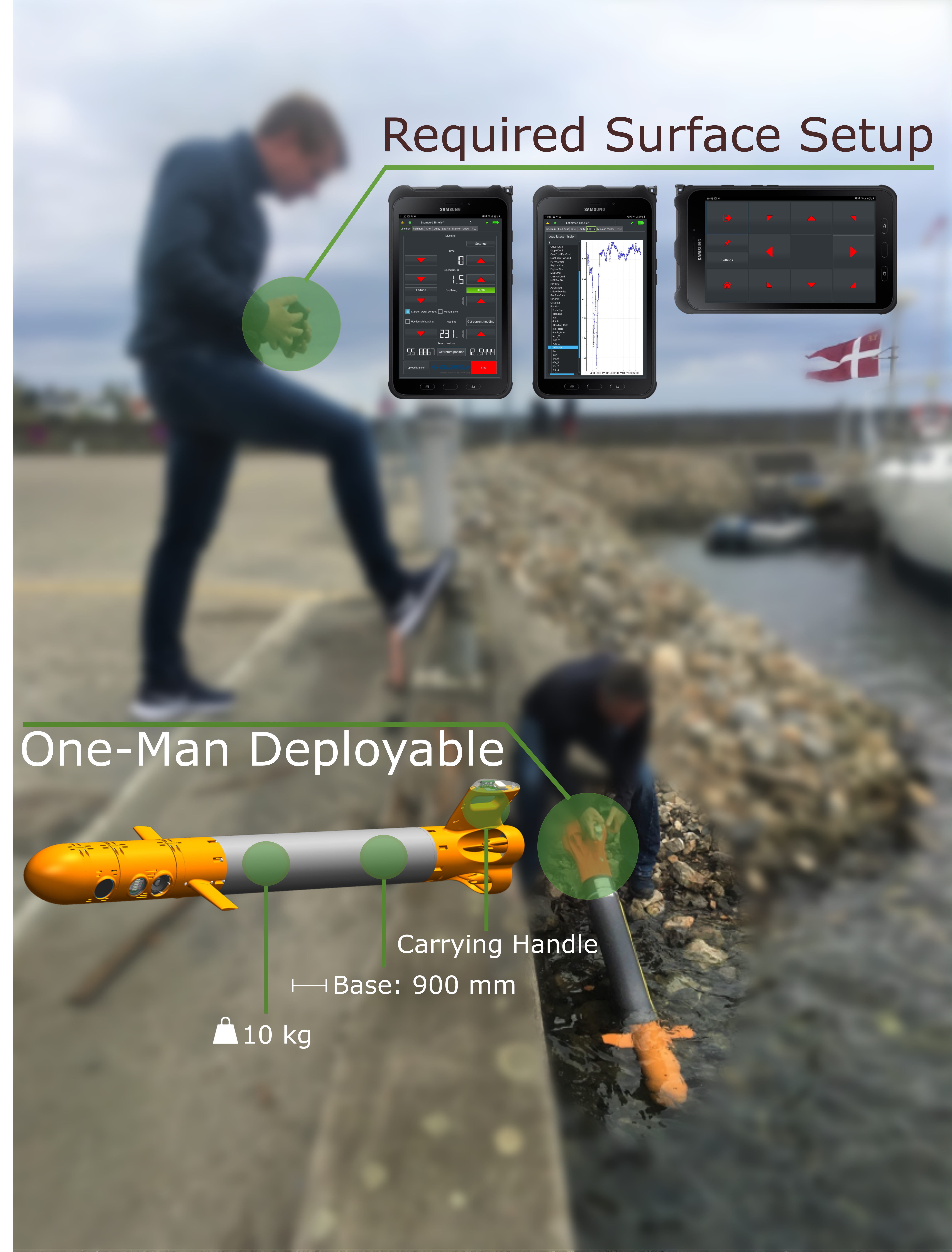}
    \caption{\textit{SeaShark AUV being deployed at a harbor.} Only one person is essential for SeaShark AUV operations. With carrying handle, small size, and low weight, it is easy to deploy and recover. For mission planning, remote control, live view, and mission review, only a tablet is required.}
    \label{fig:harbor}
\end{figure}

As we demonstrate in Fig.~\ref{fig:harbor}, SeaShark is one-man deployable. A base length of \SI{900}{\milli\metre} and a total weight of \SI{10}{\kilo\gram}, makes it easy to deploy and recover almost everywhere. We require only a tablet for operating, i.e., remote control, mission planning, live sensor view, and mission review. The typical endurance of the SeaShark AUV is experienced as sufficing for a complete days work, doing on/off missions, 
and the on-board battery management system allows us to charge the vehicle using a single cable without disassembling any modules.


\subsection{Principle of operation}
In our mission planner (on the tablet), the user can construct and configure missions suitable for many operations. The three most standard types are \textit{line missions}, \textit{site missions} and \textit{circle missions}. All of them can be manually configured and adjusted on-site in seconds before launching the vehicle. The following describes our three pre-implemented missions in more detail. 

\subsubsection{Line mission}
Provides the vehicle with a heading to follow and a depth or altitude reference. As we employ no advanced navigation other than following compass heading using basic off-the-shelf sensors, our end condition is a timeout. By the end of a successful line, the vehicle ascents to the surface, turns around, and sails back to an agreed-upon GNSS position, e.g. boat or start position. 

\subsubsection{Site mission} 
Implements the well-known lawnmower pattern. The user inputs the center of the site or area that the vehicle should cover, the amount and length of lines, and the line spacing. The mission planner then proceeds to lay out a set of lines (c.f. \textit{line missions}) and uses GNSS positions at each end of a line to calculate heading and lead-in. After a line is completed, the vehicle transits to the beginning of the next line on the surface.

\subsubsection{Circle mission}
Implements a circular pattern with some given radius and (rotational) speed for the vehicle to follow. This can be at a fixed depth or altitude or a spiral throughout the water column. This type of mission is well-suited for profiling, or as an initializer or search pattern for complete autonomy.

Applicable for all of the above is the absence of velocity measurement. Thus, any vehicle position at any given time is an estimate using start position, time, and heading as a reference without any means to account for drift or other unknown factors during mission time. 

After a mission, the vehicle re-connects to the tablet and waits until further instructions are given. Using the remote control on the tablet, the vehicle may be set to loiter on its current position if circumstances do not allow for retrieval between missions. 

\begin{figure}
    \centering
    \includegraphics[width=0.95\linewidth]{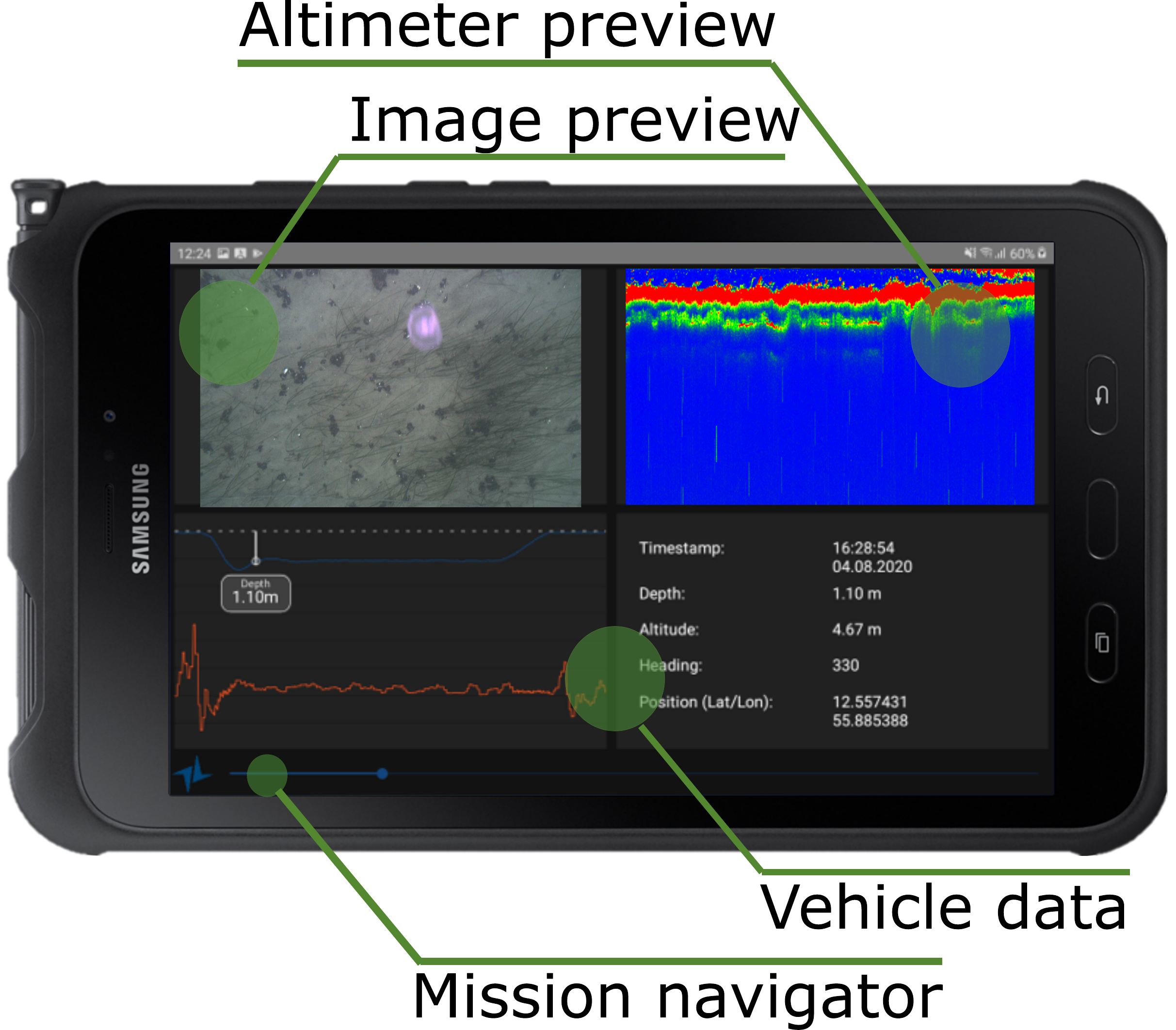}
    \caption{\textit{Post-mission preview.} The tablet is used as a quick-view or preview of the data recorded from the latest mission in the field. This quickly and easily verifies image quality and measurements during the mission. The user ``slides'' through the mission using the \textit{mission navigator} and thus has access to all necessary information to verify a successful mission.}
    \label{fig:preview}
\end{figure}

To quickly verify data and image quality of the latest mission, our tablet includes a ``quickview'' as shown in Fig.~\ref{fig:preview}. Here the user can scroll through the entire mission and quickly get an overview of image content, quality, and measurements for any given time during the mission. 

\begin{figure*}[b!]
    \centering
    \includegraphics[width=0.99\linewidth]{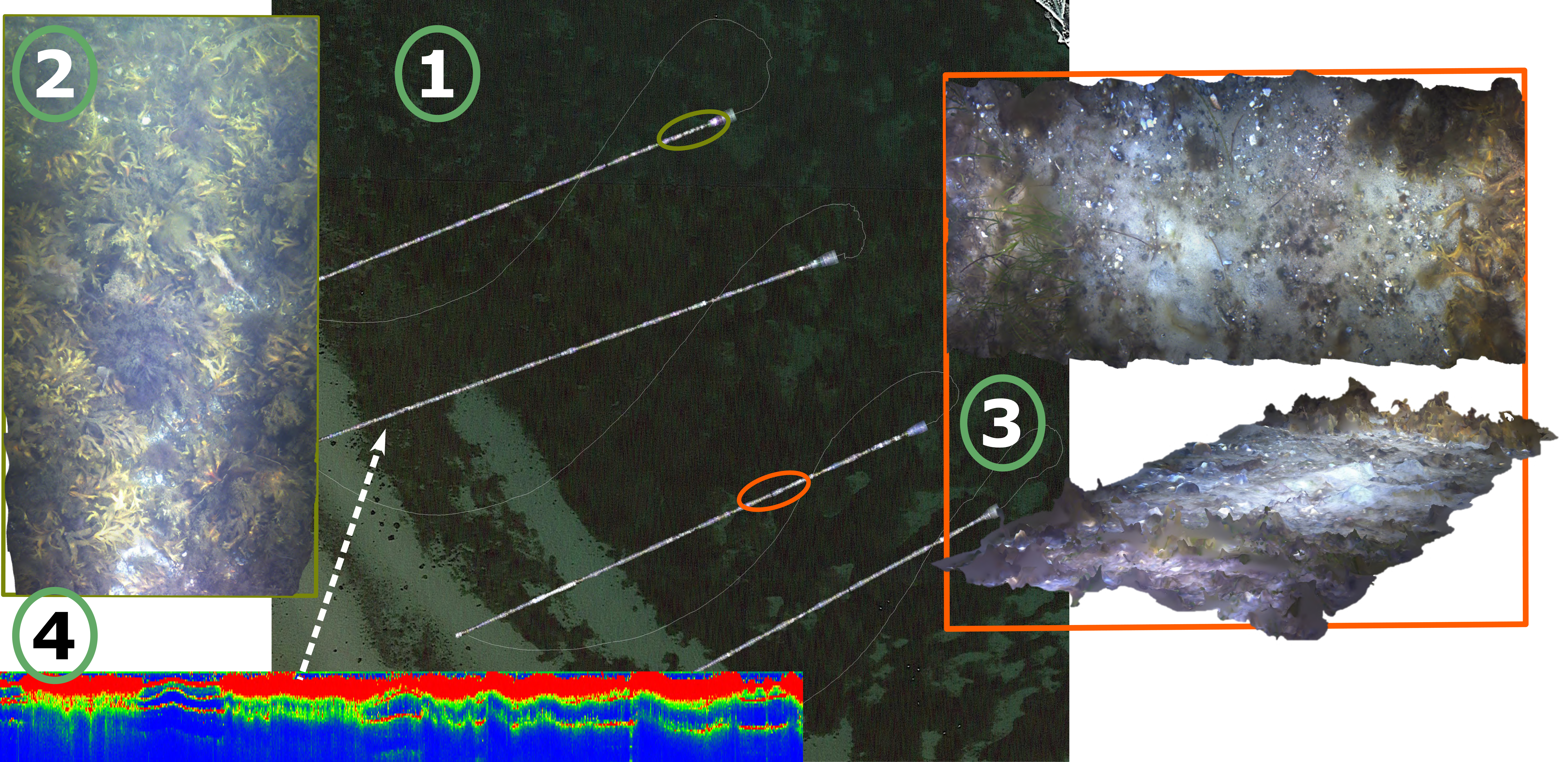}
    \caption{\textit{Case study: Eelgrass habitat.} The SeaShark AUV samples optical images and corresponding altimeter measurements in a straight line from shore to sea at a fixed altitude. \textbf{1)} Geo-referenced tracks and optical images as Google Earth overlay. \textbf{2)} Enlarged section of the image-mosaic. \textbf{3)} Enlarged section of the image-mosaic and corresponding 3D model. \textbf{4)} Altimeter measurements.}
    \label{fig:track}
\end{figure*}

\subsection{Autonomy}
The SeaShark AUV supports a high level of autonomy programmable by the user. Using a laptop, the user can configure and review event-based mission plans in addition to standard missions readily available on the tablet. Events may be triggered using any available sensor or payload on the vehicle. 

We further include a back-seat driver for navigation and autonomy. This allow users to run self-made programs, e.g. Python scripts, on the payload computer, and from this interface mission execution. This effectively enables users to script or program any algorithm or behavior by processing and acting on real-time data. For example, the vehicle may initially be running a circle mission to cover a large area for the on-board camera looking for specific objects on the seabed, e.g. ghost-nets or pipelines. When identified in the images (from the user-made program), the autonomy takes over command of the vehicle and follows or covers the object to the desired extent. After a timeout or successful coverage, the mission ends or reverts to the initial circle mission. Using the back-seat driver is as simple as broadcasting desired navigation references, i.e. heading and depth/altitude.









\section{Use-Case Examples}
Due to the small size and easily configurable nature of the SeaShark AUV, we envision the vehicle to be well suited in a broad range of tasks for both research and commercial purposes. As a result of tests and trials, we show here example data for shore-to-sea eelgrass photography and harbor wall and pier photography. 

\subsection{Eelgrass habitat}
Currently, divers and dropcams are mostly used to survey eelgrass habitats \cite{KrauseJensen2009, Dolch2017}. Using divers for such surveys is very time-consuming in preparation and execution, and require trained and licensed personal. Using dropcams limits the covered area for image acquisition to the single spots where the camera was deployed. 

Using the vehicle for such tasks may extend from shore-to-sea to deeper-sea surveys. Areas may be ``spot''-sampled using a set of parallel straight lines, or dense coverage may be achieved using site missions. With the SeaShark configured with downward-facing camera, light, and altimeter, data for eelgrass habitat inspection are easily collected. In Fig.~\ref{fig:track}, we present data and results from a sea-to-shore experiment. Here we show four tracks starting from shallow shore towards the sea. The straight lines are the actual line missions, and the S-shaped track is the transit back to the starting position on the surface. The figure further shows photo-mosaics sections, a 3D model from photogrammetry, and an altimeter water-column sample.

\subsection{Harbor wall and pier}
\begin{figure*}
    \centering
    \includegraphics[width=0.99\linewidth]{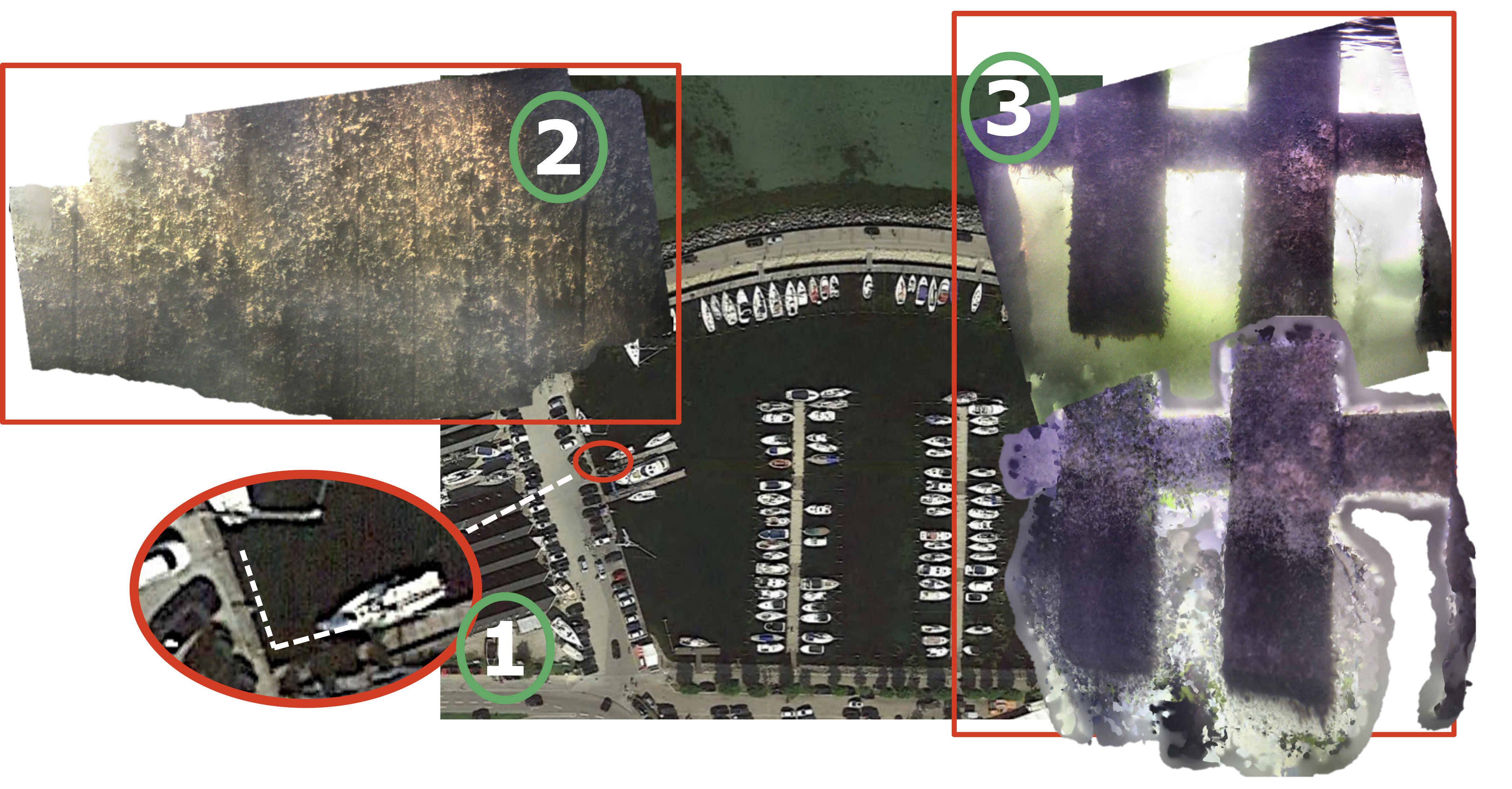}
    \caption{\textit{Case study: Harbor wall inspection.} The SeaShark AUV samples optical images along the pier and wall in a harbor. \textbf{1)} Google Earth overview and zoom of the test area. \textbf{2)} Harbor wall section as a mosaic. \textbf{3)} Harbor pier section as mosaic and 3D model.}
    \label{fig:harbor_wall}
\end{figure*}

Walls and standing structures can be covered without any modification other than changing the orientation of the camera module on the vehicle. For more complicated operations, the orientation of the altimeter or other sensor may also be modified to support navigating the vehicle (c.f. \textit{autonomy}). In Fig.~\ref{fig:harbor_wall}, we show example data of wall and pier photography at different depth levels, merged as photo mosaics. By doing so, areas can be efficiently inspected in great detail (see the crab in the down-right corner of Fig.~\ref{fig:harbor_wall}.2 as an example), and 3D models can be constructed using multiple view projections (Fig.~\ref{fig:harbor_wall}.3).

\section{Discussion and Future Work}
As previously mentioned, we rely only on the compass heading during underwater missions. Consequently, the vehicle needs to ascent to the surface each time a position reference is required. Another consequence following the absent navigation is the inability to guarantee the direction of tracks due to drift and other underwater factors. We guarantee that the vehicle follows any given heading for a fixed amount of time, but underwater conditions may cause the vehicle to drift. However, in post-processing, the GNSS-points before and after a track can be used to estimate the driven path, and accurate time-stamping then provides a rough estimate of the vehicle's position at any given time during the mission. More advanced methods, such as estimating odometry using image data, can be further applied to obtain a better estimation. 


Currently, the payload sections (the hulls) are 3D-printed. Experience shows that a consumer-grade 3D-printer and material provides adequate structural sturdiness. This provides rapid turnaround times for prototyping and allows us to iterate the design of payload sections quickly, and with minimal effort. Furthermore, producing (spare) parts for the vehicle is done quickly and at a low cost in-house. Another advantage, in the unfortunate event that we during a campaign is missing a spare part and are unable to return home, we simply need access to a 3D-printer, and hence a repair may be performed within hours. This way, the AUV is a flexible tool for both research and industrial use.

Future work will include further field testing, collaboration with universities on several topics, possible extension of payload sensors to include CTD-sensors, side-scan sonar, and acoustic aiding. 

\section{Conclusion}
This paper has introduced and presented the SeaShark AUV; A modular multi-purpose man-portable {\textmu}-AUV, which is extremely simple to operate. The entire user interaction, including planning, monitoring, and review, is handled via an intuitive, minimum-number-of-buttons GUI using a tablet.

The underwater vehicle itself is compact and can be launched and retrieved using just one hand (while holding the tablet in the other). It is conceived as modular parts that fit around a central main tube, which holds battery and other vital parts. Currently, thruster(s) are mounted at the rear, and payload sections are positioned at the bow, but this is not universally ``fixed'' in any way.  The payloads are exchangeable, stackable, and 360{\textdegree} rotatable, making the SeaShark AUV well suited for a broad range of tasks. The limitation to having GNSS only while at the surface - and using dead reckoning while submerged, has fostered a new way of conceiving and planning missions. Moreover, instead of being a limitation, this mission language has revealed itself as powerful, robust, and intuitive to such a degree that new users grasp the idea and find themselves enthusiastically \emph{using} and \emph{operating} the vehicle within minutes. 

\addtolength{\textheight}{-16cm}   

There are use cases and mission types that profit significantly from a low-complexity, light, and easy to use AUV platform. As examples of this,  we have presented data of near-shore bio-habitat mapping (eelgrass) and harbor wall and pier inspection missions. Furthermore, using a highly capable post-processing tool-set, mission data is reduced to high-quality and easy-to-interpret data suitable for on-site quick-review and detailed analysis back in the office.   

We continue our efforts to develop the AUV system. Its capabilities make it both a strong competitor, alternative and addition to towed systems, labor-intensive ROVs, and survey-grade AUVs for suitable tasks.


\bibliographystyle{bib/IEEEtran}
\bibliography{bib/IEEEabrv,bib/eelgrass}

\end{document}